\documentclass[runningheads]{llncs}
\usepackage{graphicx}

\usepackage{tikz}
\usepackage{comment}
\usepackage{amsmath,amssymb} %
\usepackage{color}
\usepackage{orcidlink}

\usepackage{booktabs}
\usepackage{multirow}
\usepackage{subcaption}
\usepackage{xspace}
\usepackage{colortbl}
\definecolor{Gray}{gray}{0.9}

\usepackage{pifont}
\newcommand{\cmark}{\ding{51}\xspace}%
\newcommand{\xmark}{\ding{55}\xspace}%

\def \testra {TeSTra\xspace}

\usepackage[accsupp]{axessibility}  %

\newcommand{\myparagraph}[1]{\vspace{0pt}\noindent{\bf #1}}
\mathchardef\mhyphen="2D
\newcommand{\etal}{\textit{et al.}\xspace}
\newcommand{\ie}{\textit{i.e.}\xspace}

\newcommand{\cR}{\mathcal{R}}

\newcommand{\vf}{\mathbf{f}}
\newcommand{\vk}{\mathbf{k}}

\newcommand{\vq}{\mathbf{q}}
\newcommand{\vs}{\mathbf{s}}
\newcommand{\vv}{\mathbf{v}}
\newcommand{\vx}{\mathbf{x}}

\newcommand{\vlambda}{\boldsymbol{\lambda}}
\newcommand{\vphi}{\boldsymbol{\phi}}
\newcommand{\vpsi}{\boldsymbol{\psi}}

\newcommand{\mK}{\mathbf{K}}
\newcommand{\mM}{\mathbf{M}}
\newcommand{\mO}{\mathbf{O}}
\newcommand{\mQ}{\mathbf{Q}}

\newcommand{\mV}{\mathbf{V}}
\newcommand{\mW}{\mathbf{W}}
\newcommand{\mX}{\mathbf{X}}
\newcommand{\mZ}{\mathbf{Z}}

\usepackage{stmaryrd}

\newcommand{\softmax}{\mathrm{Softmax}}
\newcommand{\attention}{\mathrm{Attention}}
\newcommand{\concat}{\mathbin\Vert}

\begin{document}
\pagestyle{headings}
\mainmatter
\def\ECCVSubNumber{1055}  %

\title{Real-time Online Video Detection with Temporal Smoothing Transformers} %

\titlerunning{Real-time Online Video Detection with Temporal Smoothing Transformers}
\author{Yue Zhao\inst{1}\orcidlink{0000-0003-2753-5921} \and
Philipp Kr\"ahenb\"uhl\inst{1}\orcidlink{0000-0002-9846-4369}}
\authorrunning{Y. Zhao and P. Kr\"ahenb\"uhl}
\institute{University of Texas at Austin, Austin TX 78712, USA \\
\email{\{yzhao,philkr\}@cs.utexas.edu}\\
}
\maketitle

\begin{abstract}
Streaming video recognition reasons about objects and their actions in every frame of a video.
A good streaming recognition model captures both long-term dynamics and short-term changes of video.
Unfortunately, in most existing methods, the computational complexity grows linearly or quadratically with the length of the considered dynamics.
This issue is particularly pronounced in transformer-based architectures.
To address this issue, we reformulate the cross-attention in a video transformer through the lens of kernel and apply two kinds of temporal smoothing kernel: A box kernel or a Laplace kernel.
The resulting streaming attention reuses much of the computation from frame to frame, and only requires a constant time update each frame.
Based on this idea, we build \testra, a Temporal Smoothing Transformer, that takes in arbitrarily long inputs with constant caching and computing overhead.
Specifically, it runs $6\times$ faster than equivalent sliding-window based transformers with 2,048 frames in a streaming setting.
Furthermore, thanks to the increased temporal span, \testra achieves state-of-the-art results on THUMOS'14 and EPIC-Kitchen-100, two standard online action detection and action anticipation datasets.
A real-time version of \testra outperforms all but one prior approaches on the THUMOS'14 dataset.

\end{abstract}

\section{Introduction}
The problem of online action detection~\cite{degeest2016online} and anticipation~\cite{kitani2012activity} aims to determine what action is happening or will happen shortly at each time step without seeing the future.
The challenge for online action detection is (1) how to effectively retain both the long-term trends and short-term cues when encoding the history and (2) how to efficiently compute at each time step in the streaming setting when the history gets longer.
Recurrent models such as LSTM~\cite{hochreiter1997lstm} and GRU~\cite{chung2014gru} excel at updating the output recurrently but do not benefit from increasing sequence length due to the training difficulty~\cite{werbos1990bptt}.
Attention-based models~\cite{vaswani2017attention}, like Long Short-Term Transformer (LSTR)~\cite{xu2021lstr}, are capable of handling sequences up to 8 minutes long with impressive prediction results.
However, in the streaming setting, the attention computation of the long-term memory has to be recomputed for each streaming window considered.
Therefore, the computational cost per frame is proportional to the sequence length.

\begin{figure}[t]
    \centering
    \includegraphics[width=\textwidth]{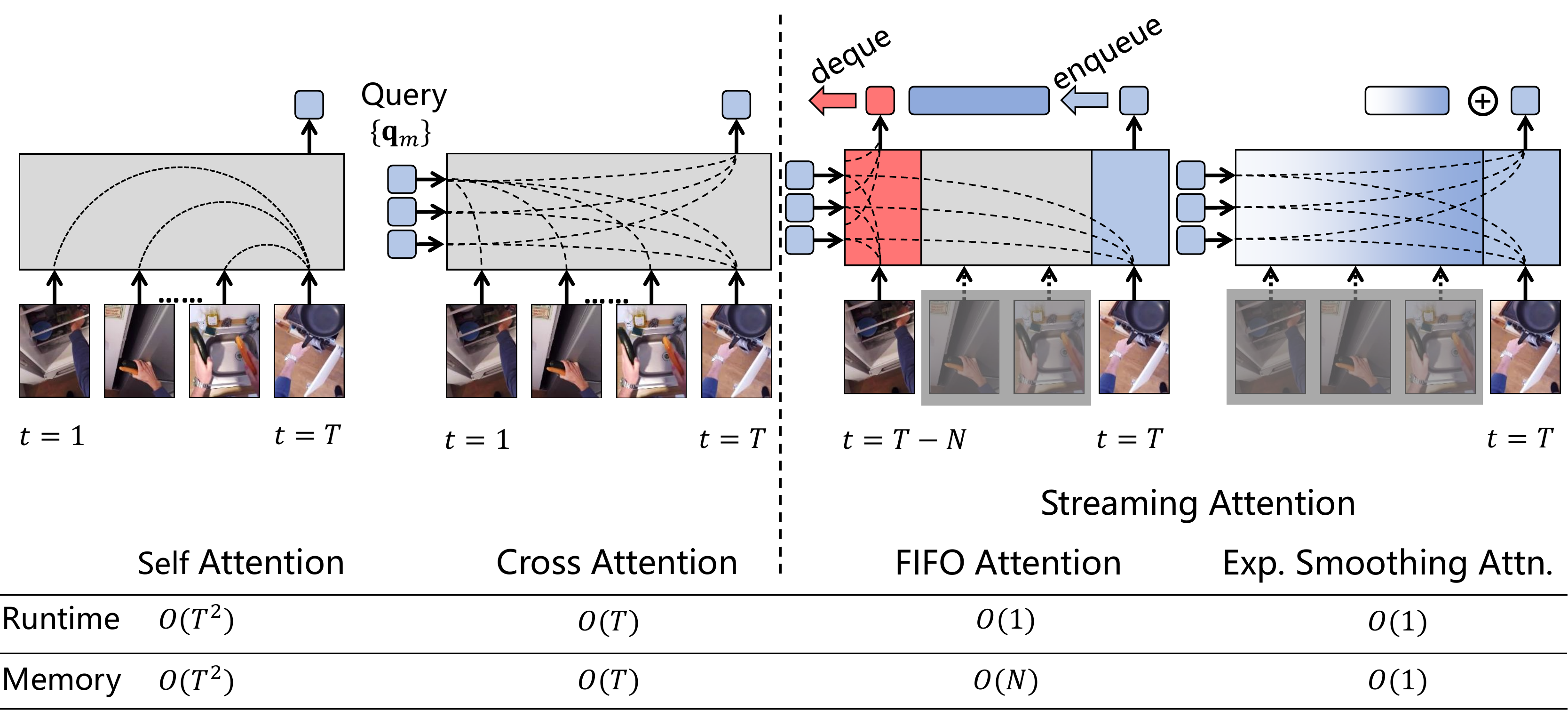}
    \caption{\small{A comparison of traditional attention computation (left) in streaming videos and our streaming attention (right).
    Unlike traditional approaches, our approach has a constant runtime per frame. Exponential smoothing attention has a constant memory footprint as well}}
    \label{fig:teaser}
    \vspace{-2mm}
\end{figure}

In this paper, we propose an effective and efficient approach, {\em Temporal Smoothing Transformers} (\testra), to encode sufficiently long history with constant inference cost at each time step.
\testra relies on an efficient attention that reuses much of the attention computation between consecutive frames.
We reformulate attention through a kernel perspective~\cite{scholkopf2002learning,tsai2019transformer} and explore two temporal kernels: a Box kernel and a Laplace kernel.
Both kernels lead to an efficient streaming attention computation.
A box kernel results in a First In First Out (FIFO) attention computation with a constant runtime update, but linear memory costs.
A Laplace kernel results in an exponential smoothing attention with constant runtime and memory costs.
Fig.~\ref{fig:teaser} shows a comparison of traditional attention for streaming videos and our streaming attention.
Both formulations exploit the fact that in streaming recognition queries used in cross attention are learned parameters and fixed during inference.
During training, we use windowed attention in its original matrix multiplication form (with explicitly computed kernels).
This allows us to enjoy all the GPU parallelism of modern transformer training.
At test time, we switch to efficient streaming implementations.

To show the effectiveness of \testra, we conduct extensive experiments on standard benchmarks for online action detection and anticipation, namely THUMOS'14~\cite{Idrees2016thumos} and EPIC-Kitchen-100~\cite{Damen2021Rescaling}.
\testra achieves state-of-the-art performance on both benchmarks.
Running at 142.8 FPS alone, \testra can serve as a building block for streaming video recognition with low latency.
When we include an accelerated optical flow computing method and an image-based feature extractor, the overall system can run as fast as 41.1 FPS and achieves 67.3\% mAP on THUMOS'14, outperforming all but one prior approaches.
Code is publicly available at \url{https://github.com/zhaoyue-zephyrus/TeSTra/}.

\section{Related Work}

\myparagraph{Online Action Detection and Anticipation.}
Online action detection~\cite{degeest2016online}, also known as early action detection~\cite{hoai2014max}, aims to detect the start of an action in a video stream as soon as it happens.
Much of prior work builds ever longer-term temporal reasoning using various recurrent units or networks~\cite{degeest2018modeling,eun2020idn,xu2019trn}.
Xu~\etal~\cite{xu2019trn} perform online detection (classification) on current frame and prediction the near-future actions simultaneously.
StartNet~\cite{gao2019startnet} decomposes the online detection into two stages: action classification and start localization.
The recently proposed LSTR~\cite{xu2021lstr} enlarges the effective temporal context to as long as 512 seconds by adopting the highly flexible cross-attention mechanism in Transformer~\cite{vaswani2017attention}.
However, the induced computation cost is proportional to the temporal span.
In contrast, our streaming attention incurs the same constant runtime cost independent of temporal span.

Action anticipation~\cite{girdhar2021avt}, or forecasting~\cite{kitani2012activity}, aims to predict the action {\em before} it occurs.
Vondrick~\etal~\cite{vondrick2016anticipating} propose to anticipate by regressing the representations of future frames from past ones.
Zeng \etal~\cite{zeng2017visual} and Rhinehart \etal~\cite{rhinehart2017first} use inverse reinforcement learning to perform forecasting at multiple levels.
For egocentric videos anticipation may additionally incorporate the camera wearer's trajectory~\cite{park2016egocentric}, eye gaze~\cite{li2018eye}, hand-object interaction~\cite{koppula2013learning}, and environment affordance~\cite{nagarajan2020egotopo}.
In this paper, we handle the problem by taking longer history into account, which is a general approach to both third-person and egocentric videos.

\myparagraph{Transformers and its Efficient Variants.}
Since the Transformer architecture was introduced in~\cite{vaswani2017attention}, much work has gone into improving the efficiency of dot-product attention.
Low-rank approximation on attention matrix~\cite{choromanski2021performer,wang2020linformer} factorizes the attention matrix into two lower-rank matrices.
Different efficient learnable sparsity patterns, such as locality-sensitive hashing~\cite{kitaev2020reformer}, differentiable sorting~\cite{tay2020sparse} or fixed patterns~\cite{child2019generating,zaheer2020bigbird}, reduce the total number of attention operations.
Query-based cross attention mechanisms compress longer-term input into a fixed-size representation via memory~\cite{rae2020compressive,lei2020mart} or recurrence~\cite{dai2019transformerxl}.
Based on a kernel-reformation~\cite{tsai2019transformer}, Katharopoulos~\etal~\cite{katharopoulos2020transformers} propose linear attention by decomposing the kernel function $\kappa(\vq_m, \vk_n)$ between a query-key pair into a product between the feature mapping of query and key,~\ie $\phi(\vq_m)^\top\cdot\phi(\vk_n)$.
In computer vision, Transformers are made more efficient by
(1) leveraging hierarchy using shifted local window~\cite{liu2021swin} and pooling attention~\cite{fan2021mvit},
(2) applying axial attention on separate dimensions~\cite{wang2020axial},
and (3) using asymmetric attention (cross attention) to squeeze high-dimensional inputs into tighter latent variables~\cite{jaegle2021perceiver}.
In speech recognition, transformers are tailored to streaming decoding by integrating recurrence~\cite{zhang2020transformer} or memory~\cite{wu2020streaming}.
In this paper, we follow the kernel interpretation of Tsai~\etal~\cite{tsai2019transformer}, and show how to efficiently update streaming attention kernels.

\myparagraph{Efficient Video Processing.}
Videos are notoriously expensive to process.
TSN~\cite{wang2018tsn} suggests sampling frames sparsely and running 2D CNNs on the selected frames.
MVCNN~\cite{zhang2016mvcnn} and CoViAR~\cite{wu2018coviar} directly learn video representation from compressed videos.
X3D~\cite{feichtenhofer2020x3d} and CSN~\cite{tran2019csn} reduce computation FLOPs by leveraging channel-wise separable convolution.
However, 3D CNN takes video clips as input whose span can be $2-3$ seconds, therefore may not be the best solution in a low-latency application.
Par-Inception~\cite{carreira2018massively} tackles the latency issue by introducing depth-parallelism to the vanilla I3D~\cite{carreira2017kinetics} at increased implementation difficulty.
Most of the previous methods focus on trimmed videos whose duration is often in several seconds while our method focuses on streaming videos whose length can be as long as hours.
However, many of these 3D CNNs may form a good backbone to our system.

\section{Preliminaries}

\myparagraph{Attention.}
The attention mechanism~\cite{vaswani2017attention} is a weighted addition of the input features.
The weights are guided by the similarities between the key and query elements on an input sequence:
\begin{align}
    \label{eq:attention}
    \attention(\mQ, \mX)
    = \softmax\left(\frac{\mQ \mK^\top}{\sqrt{C}}\right)\cdot\mV
    = \softmax\left( \frac{\mQ\cdot(\mX\mW_k)^\top}{\sqrt{C}} \right)\cdot \mX\mW_v,
\end{align}
where $\mQ\in\cR^{M\times C}$ is a set of $M$ queries, $\mX=\begin{bmatrix} \cdots & \vx_n & \cdots \end{bmatrix}^\top\in\cR^{N\times d}$ is the sequence of $N$ input tokens, $\mW_{k/v}\in\cR^{d\times C}$ is the weight to map the input to key/value vector and $C$ is the feature dimension of $\vx_n^\top\mW_k$.
For self-attention computes queries from the inputs sequence $\mQ=\mX\mW_q$ ($M=N$ in this case).
Cross-attention uses a queries $\mQ$ that do not relate to the input sequence $\mX$ (generally $M \ne N$ in this case).
Cross-attention is commonly used in the encoder-decoder architecture~\cite{vaswani2017attention}.
Cross-attention with $M \ll N$ is also used to efficiently encode large amounts of data into a fixed-size representation~\cite{jaegle2021perceiver,xu2021lstr}.

\myparagraph{Attention as kernels.}
The distance computation in attention is similar to the mechanism of kernel learning~\cite{scholkopf2002learning}.
Tsai~\etal\cite{tsai2019transformer} reformulated Eq.~\eqref{eq:attention} from the perspective of kernels:
\begin{align}
    \label{eq:attention_kernel}
    \attention(\vq_m,\{\vx_n\})
    = \frac{\sum_{n=1}^N \kappa(\vq_m,\vk_n)\vv_n }{\sum_{n=1}^N \kappa(\vq_m,\vk_n)},
\end{align}
where $\kappa(\cdot,\cdot): \cR^{C}\times\cR^{C}\rightarrow\cR^+$ is a generalized kernel function, which depicts the similarity between the pair of input vectors.
Eq.~\eqref{eq:attention} is equivalent to Eq.~\eqref{eq:attention_kernel} for a kernel $\kappa(\vq_m, \vk_n)=\exp(\frac{\vq_m^\top\vk_n}{\sqrt{C}})$.
In the next section, we show that this kernel perspective leads to an efficient streaming formulation of attention in the context of streaming video recognition.

\section{Efficient Attention on streaming input}

We use cross-attention to summarize a large stream of past frames into a fixed size context representation.
We use a fixed number learned queries and variable number of keys and values from past frames as input.
See Fig.~\ref{fig:teaser} for an example.

In streaming tasks, we are constantly receiving input and want to generate the corresponding output on the fly.
Examples include simultaneous interpretation or online detection in broadcast videos.
Let $\vx_{[1:t]} = \{ \vx_1, \vx_2, \cdots, \vx_t \}$ denote a sequence of encoded past video frames for the current time-step $t$.
The encoder may use an image-based ~\cite{he2016resnet,ioffe2015batchnorm} or short-clip-based~\cite{carreira2017kinetics,feichtenhofer2020x3d} CNN.
Top-performing video models~\cite{xu2021lstr} summarize large parts of the video through cross-attention on either the entire sequence,~\ie $\attention(\vq_1\ldots\vq_M,\vx_{[1:t]})$ or a chunk of input by sliding a $N$-sized temporal window,~\ie $\attention(\vq_1\ldots\vq_M,\vx_{[t-N+1:t]})$.
Mathematically, this attention operation is captured in Eq.~\eqref{eq:attention_kernel}.
Here, a small number of queries $\{ \vq_1\ldots\vq_M \} $ summarize a large temporal context.
Queries combine learned parameters $\{ \vlambda_1\ldots\vlambda_M \}$ with a temporal embedding $\omega_t$ of the current frame: $\vq_m = \vlambda_m + \omega_t$.
Keys $\{ \vk_1\ldots\vk_t \}$ combine a frame-level embeddings $\vf_n = \mW_k^\top \vx_n$ with a temporal embedding $\omega_t$: $\vk_n = \vf_n + \omega_n$.
Values $\{ \vv_1\ldots\vv_t \}$ use the same frame-level features $\vv_n = \mW_v^\top \vx_n$.
In this setup, keys and values of past frames remain unchanged, learned queries are constant during inference, only the temporal query embedding changes frame to frame.
This changing temporal embedding does change the attention kernel $\kappa$ for each new frame.
This means in a streaming setting, we have no choice but to recompute the entire attention operation frame after frame.
This recomputation grows linearly with the size $N$ of the temporal context considered.
Next, we show how a reformulation of the attention mechanism leads to a much more efficient streaming evaluation.

\begin{figure}[t]
    \centering
    \begin{subfigure}[b]{0.48\textwidth}
    \centering
    \includegraphics[width=\textwidth]{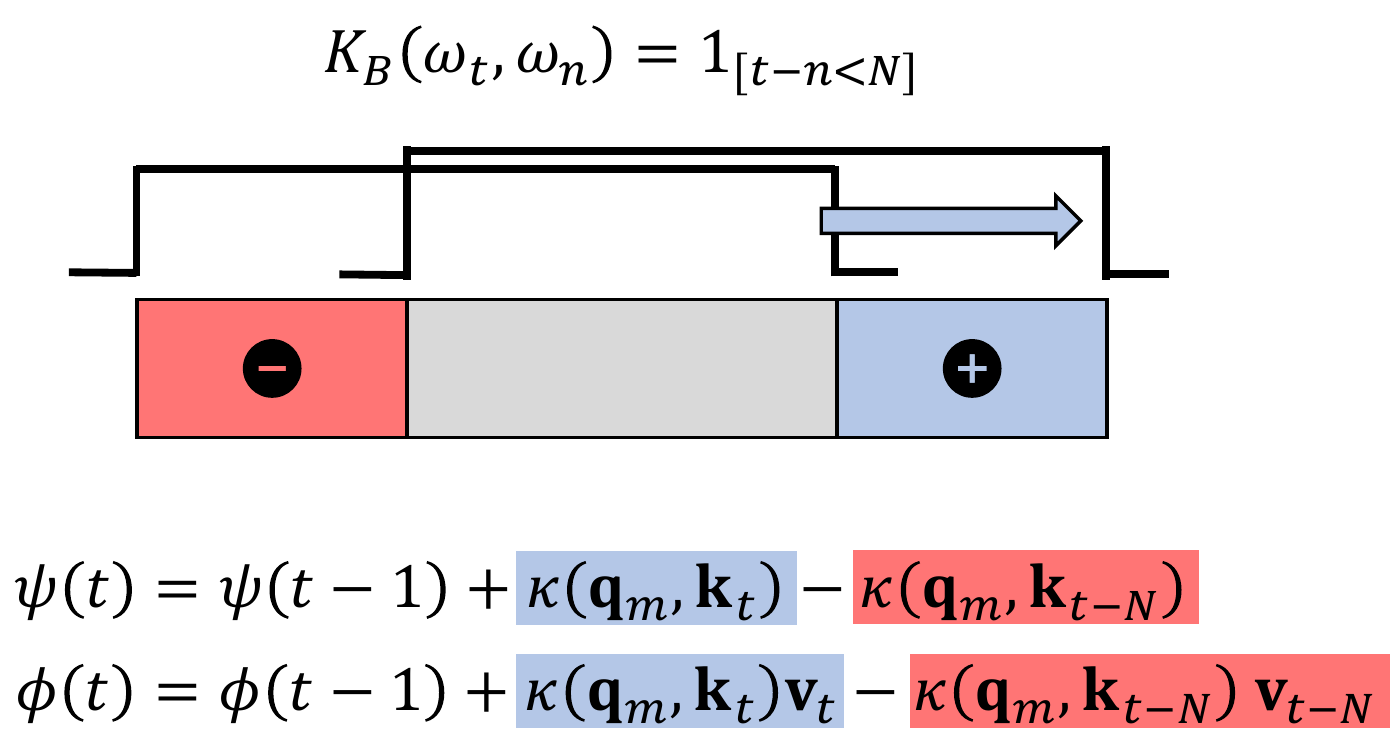}
    \caption{\small{Box kernel}}
    \label{fig:kernel_box}
    \end{subfigure}
    \hfill
    \begin{subfigure}[b]{0.48\textwidth}
    \centering
    \includegraphics[width=\textwidth]{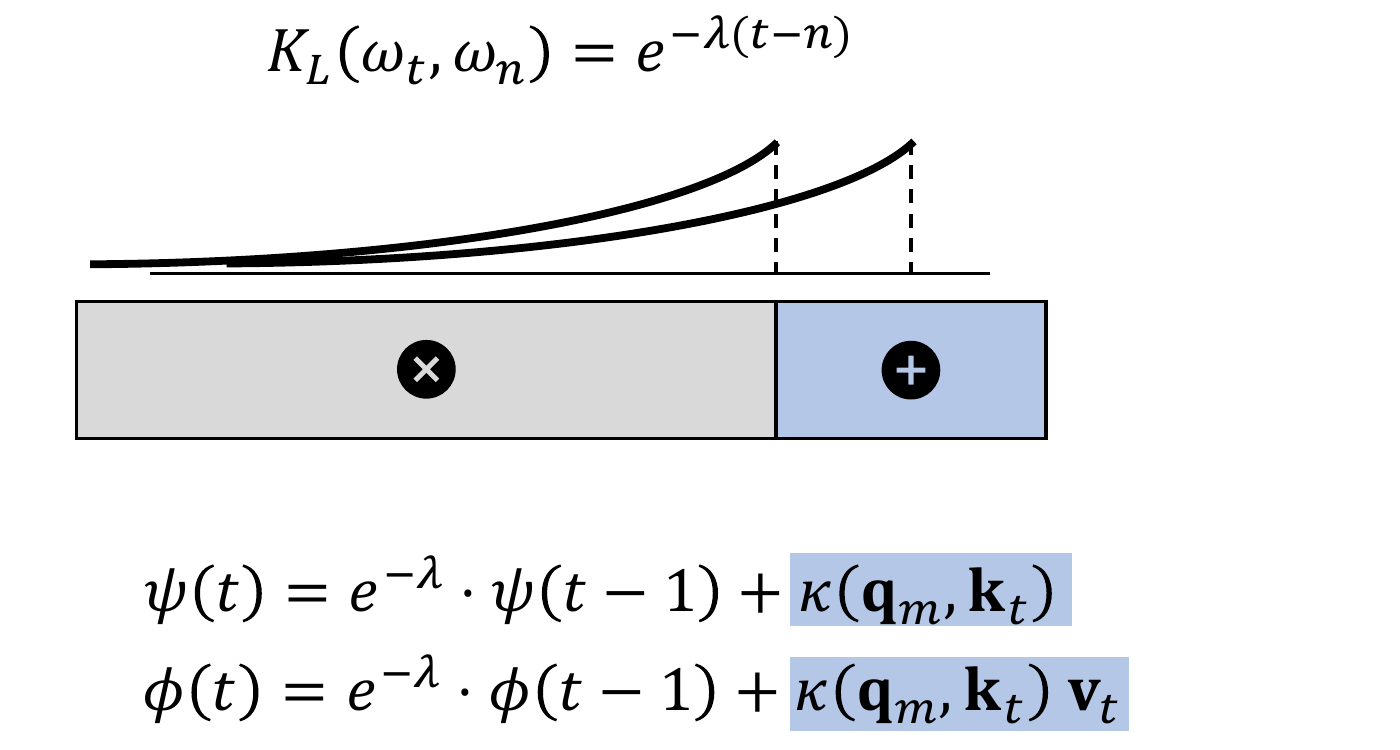}
    \caption{\small{Laplace kernel}}
    \label{fig:kernel_laplace}
    \end{subfigure}
    \caption{\small{A visualization of a box kernel (a) and Laplace kernel (b) and their streaming computation}}
    \label{fig:temporal_kernel}
    \vspace{-2mm}
\end{figure}

\myparagraph{Streaming Attention.}
Note, that both queries and keys combine a temporal and feature-level embedding in their distance kernel $\kappa(\vq_m, \vk_n) = \kappa(\vlambda_m + \omega_t, \vf_n + \omega_n)$.
In {\em Streaming Attention}, we simple split this kernel into temporal and feature component:
$K(\omega_t, \omega_n)\kappa(\vlambda_m,\vf_n)$.
The {\em Streaming Attention} operation reduces to
\begin{align}
    \label{eq:stream_attn}
    \mathrm{Stream}\mhyphen\attention(\vq_m, \vx_{[1:t]})
    = \frac{\sum_{n=1}^t K(\omega_t,\omega_n)\kappa(\vlambda_m,\vf_n)\vv_n }{\sum_{n=1}^t K(\omega_t,\omega_n)\kappa(\vlambda_m,\vf_n)}.
\end{align}
Most of the features and kernels used in this attention block remain constant throughout the streaming setting.
Moving from timestep $t$ to $t+1$ only changes the temporal kernel $K(\omega_t, \omega_n)$ to $K(\omega_{t+1}, \omega_n)$ and adds one more element $(\vf_{t+1}, \vv_{t+1})$.
Because of the change in the temporal kernel, a naive evaluation of streaming attention \eqref{eq:stream_attn} still requires a linear runtime in the size of the temporal context.
However, the right choice of a temporal kernel can alleviate this.
Here, we explore two kernels: A box (or uniform) kernel $K_B(\omega_t, \omega_n) = 1_{[t-n<N]}$ and a Laplace kernel $K_L(\omega_t, \omega_n) = \exp(-\lambda(t-n))$ for $\lambda > 0$.
Each of these kernels leads to an efficient streaming attention mechanism.
A box kernel results in first-in-first-out (FIFO) attentionw while a Laplace kernel leads to exponential smoothing attention.
Fig.~\ref{fig:temporal_kernel} provides an overview of both kernels.

\myparagraph{FIFO Attention.}
Let us define the numerator and denominator of Eq.~\eqref{eq:stream_attn} to be two intermediate variables
\begin{align}
    \mathrm{Stream}\mhyphen\attention(\vq_m, \vx_{[1:t]})=\frac{\vphi(t)}{\vpsi(t)}.
\end{align}
Both $\vphi(t)\!=\!\sum_{n=1}^t\!K_B(\omega_t,\omega_n)\kappa(\vlambda_m,\vf_n)\vv_n$ and $\vpsi(t)\!=\!\sum_{n=1}^t\!K_B(\omega_t,\omega_n)\kappa(\vlambda_m,\vf_n)$ are updated by the following recursion as the streaming attention progresses:
\begin{align}
    \label{eq:fifo_attn_recursion}
\begin{split}
    \vphi(t+1) &= \vphi(t) + \kappa(\vlambda_m,\vf_{t})\vv_{t} - \kappa(\vlambda_m,\vf_{t-N})\vv_{t-N}\\
    \vpsi(t+1) &= \vpsi(t) + \kappa(\vlambda_m,\vf_{t})\vv_{t} - \kappa(\vlambda_m,\vf_{t-N}),
\end{split}
\end{align}
where $\kappa(\vlambda_m,\vf_{t-N}) = 0$ and $\vv_{t-N} = 0$ for $t \le N$,  $\vphi(0)=0$ and $\vpsi(0)=0$.

Like a FIFO queue, we keep track of $\vphi(t)$ and $\vpsi(t)$ and update them by subtracting the quantity contributed by the input at time $(t-N)$ and adding up the one at time $t$ in the long run.
Therefore, we call this formulation {\em FIFO-Attention}.
The advantage of FIFO-Attention is that the computational cost becomes $O(MC)$ for $M$ queries and values of $C$ channels.
Neither the effective window size $N$ nor the actual time-step $t$ influences the runtime.
However, the subtraction operation in Eq.~\eqref{eq:fifo_attn_recursion} requires us to keep a window of features and kernel values in memory.
Hence, the memory complexity is still $O(N)$.
The Laplace kernel addresses this issue.

\myparagraph{Exponential Smoothing Attention.}
The Laplace kernel $K_L$ allows for an even more efficient recursive update:
\begin{align}
    \label{eq:es_attn_recursion}
\begin{split}
    \tilde{\vphi}(t) &= e^{-\lambda}\tilde{\vphi}(t-1) + \kappa(\vlambda_m, \vf_t)\vv_t\\
    \tilde{\vpsi}(t) &= e^{-\lambda}\tilde{\vpsi}(t-1) + \kappa(\vlambda_m, \vf_t),
\end{split}
\end{align}
where $\tilde{\vphi}(0)=0$ and $\tilde{\vpsi}(0)=0$.
The parameters $\lambda$ controls the temporal extent of the attention.
The above operation \eqref{eq:es_attn_recursion} is known as exponential smoothing~\cite{holt2004forecasting}.
Therefore we name this attention {\em Exponential Smoothing Attention}, or {\em ES-Attention} for short.
Both ES- and FIFO-Attention reduce to the same operation if $\lambda=0$ and the windows size $N\to\infty$.
The time complexity of ES-Attention is also constant in the temporal window considered $O(MC)$.
More importantly, the space complexity reduces from $O(N)$ to $O(1)$ since we only maintain $\tilde\vpsi$, $\tilde\vphi$ and no longer keep values in our window around.
Exponential smoothing instead slowly reduces the influence of older keys and values in the attention.

\begin{figure}[t]
    \centering
    \includegraphics[width=\textwidth]{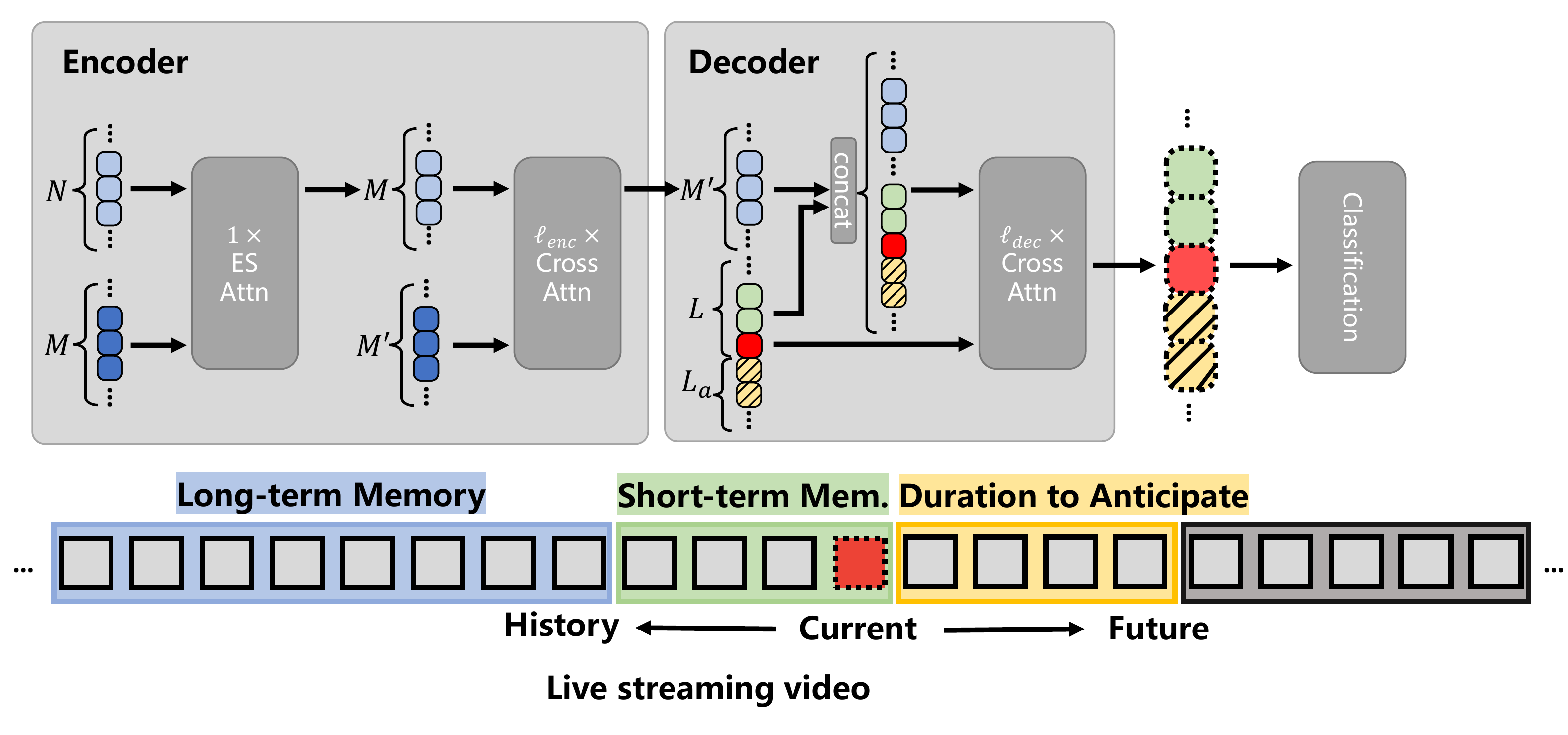}
    \caption{\small{
    Overview of our streaming attention architecture \testra.
    The basic setup follows LSTR~\cite{xu2021lstr}:
    A long-term memory compresses a long temporal history into $M$ representative queries.
    A short-term attention mechanism uses the compressed memory and a short history of frames to compute current and future actions.
    The main advantage of \testra is that the long-memory incurs only constant cost, and thus allows for much more efficient long-term reasoning}}
    \label{fig:pipeline}
    \vspace{-2mm}
\end{figure}

\myparagraph{Video recognition with streaming attention.}
The streaming attention can replace the vanilla cross-attention in current Transformer architectures with minimal modification.
Specifically, we follow LSTR architecture~\cite{xu2021lstr} for all our experiments, due to its state-of-the-art performance on online action detection.
The overall architecture of \testra is sketched in Fig.~\ref{fig:pipeline}.
Given a sequence of encoded vectors $\vx_{[1:t]} = \{ \vx_1, \vx_2, \cdots, \vx_t \}$, where $t$ refers to the current time stamp, we divide the historic frames into two parts: short-term memory $\vx_{[t-L+1:t]}$ if size $L \le 32$ and long-term memory which contains the rest of distant inputs, namely $ \vx_{[1:t-L]}$.
The architecture follows an encoder-decoder~\cite{vaswani2017attention,xu2021lstr} design.
The encoder module encodes the long-term memory into $M=16$ query features.
The decoder uses the query features and short-term memory to predict current and anticipated actions.

\begin{figure}[t]
    \centering
    \includegraphics[width=\textwidth]{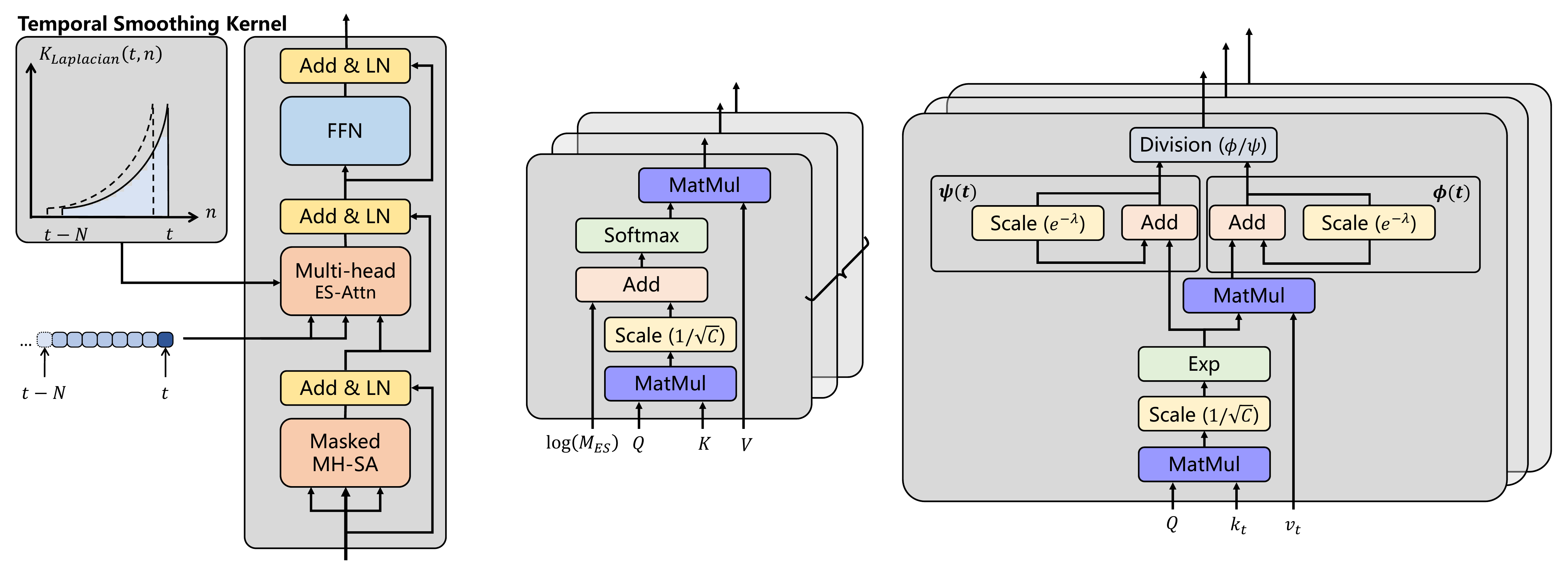}
    \caption{\small{The basic building blocks of \testra. Left: the Transformer Encoder with ES-Attention; Middle: Multi-head ES-Attention at training time; Right: Multi-head ES-Attention at inference time}}
    \label{fig:es_attention}
    \vspace{-2mm}
\end{figure}

The encoder has two stages of memory compression.
First, it uses an ES-Attention-based Transformer decoder unit~\cite{vaswani2017attention} to compress the long-term memory into $M$ latent vectors $\mZ$ using learnable queries $\mQ$.
\begin{align}
    \label{eq:encoder_selfattn}
\begin{split}
    \mQ' &= \attention(\mQ,\mQ) , \\
    \mZ' &= \mathrm{ES}\mhyphen\attention(\sigma(\mQ'), \mX_{[1:t-L]}) , \\
    \mZ &= \mathrm{FFN}(\sigma(\mZ')) ,
\end{split}
\end{align}
where $\sigma$ denotes the nonlinear mapping which is composed of a skip connection with $\mQ$ followed by a LayerNorm~\cite{ba2016layernorm}.
Next, the compressed vectors are further cross-attended by $M'$ learnable queries through $\ell_{enc}$ decoder units into $\mZ_{\ell_{enc}}$.
Strictly speaking, it should be possible to learn $\mQ^\prime$ directly.
However, the training dynamics of transformer work out better using a self-attention block first.
Fig.~\ref{fig:es_attention} shows an overview of the encoder.

The decoder uses the short-term memory as queries to attend the compressed memory and retrieve relevant information through a stack of $\ell_{dec}$ decoder units.
\begin{align}
    \label{eq:decoder_xattn}
\begin{split}
    \mX_{[t-L+1:t]}' &= \attention(\mX_{[t-L+1:t]}, \mX_{[t-L+1:t]}) , \\
    \mO' &= \attention(\mX_{[t-L+1:t]}', \left[\mZ_{\ell_{enc}} \concat \mX_{[t-L+1:t]} \right] ) , \\ 
    \mO &= \mathrm{FFN}(\sigma(\mO')) , 
\end{split}
\end{align}
In Eq.~\eqref{eq:decoder_xattn}, we construct the key/value tokens by concatenating $ \left[\cdot\concat\cdot\right] $ both the compressed long-term memory and short-term memory to incorporate all the known historic information.
This proves to be effective for action anticipation, where the closer memory is more important to indicate the upcoming action.
The $L$ output vectors are then passed through a linear layer to produces the scores $\vs_{[t-L+1:t]}\in\cR^{L\times (K+1)}$ over $K$ action classes plus one non-action (background) class
\footnote{$\vs_t\in\cR^{K}$ if the background class is absent.}.
At inference time, we take the score $\vs_t$ to be online detection result.
In action anticipation, the frames in the anticipating duration are not observable.
We thus attach $L_{a}$ learnable tokens after short-term memory predict $L_{a}$ anticipated actions $\vs_{[t+1:t+L_a]}$.

\myparagraph{Training \testra.}
At inference time, we naturally apply the recursion in Eq.~\eqref{eq:es_attn_recursion} in the streaming setting.
During training, however, it is computationally inefficient to feed all historic inputs and update them recursively on a modern GPU architecture.
To handle this, we cut the video into a clip $\vx_{t-L-N+1:t}$.
Multiple clips share the same length $N$ and thus can be packed into a batch.
Furthermore, instead of recursion, we compute the attention in matrix form:
\begin{align}
    \label{eq:es_attn_window}
    \mathrm{ES}\mhyphen\attention_{train}(\mQ'',\mX)
    &= \softmax\left(\log(\mM_{ES}) +  \frac{\mQ''\mK^\top}{\sqrt{C}}\right)\cdot\mV,  \\
    \mM_{ES}
    &=
    \begin{bmatrix}
    e^{-\lambda(N-1)} & e^{-\lambda(N-2)} & \cdots & 1 \\
    e^{-\lambda(N-1)} & e^{-\lambda(N-2)} & \cdots & 1 \\
    \vdots & \vdots & & \vdots \\
    e^{-\lambda(N-1)} & e^{-\lambda(N-2)} & \cdots & 1 \\
    \end{bmatrix},
\end{align}
where $\log(\cdot)$ takes the element-wise logarithm of a matrix and the exponential smoothing matrix $\mM_{ES}\in\cR^{M\times N}$ is a Vandermonde matrix.
Since we train on the windowed input and test on un-windowed streaming input, we select a decay factor $\lambda$ and window size $N$ such that $e^{-\lambda(N-1)}$ is sufficiently small.
This minimizes the effect of a potential train-test gap.
Fig.~\ref{fig:es_attention} shows the difference between training and inference for streaming attention.

We use the cross-entropy loss to predict both current and anticipated actions.
Following~\cite{xu2021lstr,girdhar2021avt}, we predict actions for all frames in short-term memory for a stronger supervisory signal.
We use a causal attention mask~\cite{vaswani2017attention} on the short-term memory to avoid future actions from influencing our predictions.

\section{Experiments}

\subsection{Experimental Setup}
\myparagraph{Datasets.}
We conduct experiments on THUMOS'14~\cite{Idrees2016thumos} and Epic-Kitchen-100 (EK100)~\cite{Damen2021Rescaling}.
THUMOS'14 contains 413 untrimmed videos annotated with 20 actions.
We train our model on the validation set (200 videos) and evaluate on the test set (213 videos).
Epic-Kitchen-100 contains 100 hours of egocentric videos with 90K action segments.
The narrations are mapped into 97 verb classes and 300 noun classes.
We follow the train/val split given by Furnari~\etal~\cite{furnari2020rulstm}.

\myparagraph{Evaluation Metrics.}
For THUMOS'14, we measure the performance of both online action detection and anticipation with per-frame mean average precision (mAP).
Anticipation mAP uses an anticipation period $\tau_o$ which varies from 0.25s to 2.0s with a stride of 0.25s.
Online detection mAP is as a special case of anticipation mAP at $\tau_o = 0$.
EK-100 uses mean Top-5 Verb/Noun/Action Recall to measure anticipation performance per instance with a predefined $\tau_o=1s$~\cite{Damen2021Rescaling}.

\myparagraph{Implementation Details.}
On THUMOS14, we pre-process the videos into 24 FPS, extract the two-stream deep features pretrained on ActivityNet or Kinetics following Xu~\etal~\cite{xu2021lstr}.
The visual stream is a ResNet-50~\cite{he2016resnet} while the motion stream uses BN-Inception~\cite{ioffe2015batchnorm}.
On EK100, we first pre-process the videos into 30 FPS and fine-tune the two-stream TSN~\cite{wang2018tsn} on EK100 action classification task with ImageNet-pretrained parameters, following Furnari~\etal~\cite{furnari2020rulstm}.
When training~\testra, we apply equalization loss~\cite{tan2020equalization} to handle the long-tailness of actions.
Our model is not restricted to using 2D CNNs as backbone.
Efficient 3D CNN such as X3D~\cite{feichtenhofer2020x3d} is also applicable but the longer input span might cause higher latency.

The training procedure of \testra on THUMOS'14 follows Xu~\etal~\cite{xu2021lstr} for fair comparison.
Specifically, we train \testra with batch size of 16 for 25 epochs using Adam optimizer with a weight decay of 5e-5 and a base learning rate of 7e-5.
We apply a cosine annealing schedule with linear warm-up, ~\ie the learning rate linearly increases from 0 to 7e-5 in the first 10 epochs and then decays following a cosine function.

\begin{figure}[t]
\centering
\includegraphics[width=0.8\textwidth]{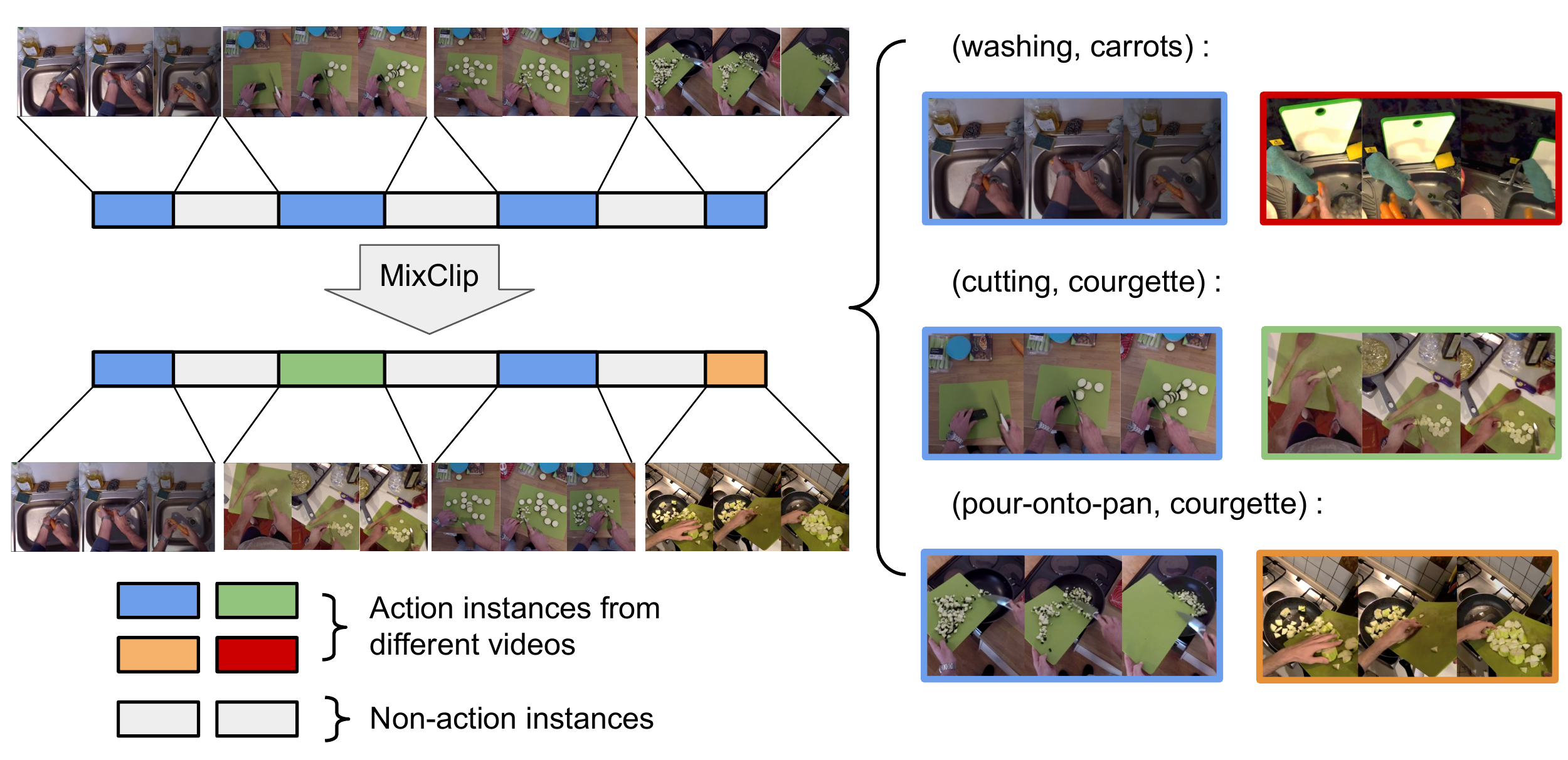}
\caption{\small{Illustration of MixClip. In the example sequence, we have 4 action instances and 2 of them are replaced by another clip that comes from another video but is annotated with the same action category}}
\label{fig:mixclip}
\vspace{-2mm}
\end{figure}

\myparagraph{MixClip.}
The model takes as input both short clips as working memory and frames in the longer history as long-term memory.
The duration of long history is significantly larger (often $10\times$) than the short clip.
This means that two neighboring clips of interest share a large portion of historical frames.
This causes the model to overfit to those scene-related cues and fail to generalize to unseen scenarios. 
To resolve this, we propose a simple augmentation technique called {\it MixClip} which increases the diversity of long history by composing short clips from different recordings into each other. 

Assume that the long memory is composed of a sequence of action instances $\{ (t_i^{(s)}, t_i^{(e)}, a_i) \}$, where $t_i^{(s)}, t_i^{(e)}$ denotes the start and end time while $a_i$ denotes the action label.
With probability $p_{mc}$, each of the action instances may be replaced with another instance with the same label from a different video.
This input feature sequence is randomly cropped if the new instance's duration longer.
Otherwise, the input feature sequence is padded to ensure that the length of history is unchanged for ease of implementation.
Fig.~\ref{fig:mixclip} gives an illustration.

MixClip is inspired by some popular augmentation techniques widely used in image classifications, such as
CutOut~\cite{devries2017cutout}, Mixup~\cite{zhang2018mixup}, and CutMix~\cite{yun2019cutmix}.

\subsection{Main Results}

\begin{table}[t]
\centering
\caption{\small{Result of online action detection on THUMOS'14.
$^\lightning$ denotes optical flow computed by NVIDIA Optical Flow SDK, a faster alternative to TV-L1~\cite{zach2007tvl1}.
More detailed runtime analysis will be provided in Sec.~\ref{sec:exp:runtime}}}
\label{table:online_detection_thumos}
\begin{subtable}[h]{0.45\linewidth}
\centering
\caption{\small{Using ANet-pretrained feature}}
\setlength{\tabcolsep}{1.0em}
\begin{tabular}{ll}
\toprule\noalign{\smallskip}
Method & mAP  \\
\noalign{\smallskip}
\cmidrule{1-2}
\noalign{\smallskip}
RED~\cite{gao2017red}       & 45.3 \\
IDN~\cite{eun2020idn}       & 50.0 \\
TRN~\cite{xu2019trn}        & 47.2 \\
OadTR~\cite{wang2021oadtr}  & 58.3 \\
LSTR~\cite{xu2021lstr}      & 65.3 \\
Ours                        &  {\bf 68.2} \\
\bottomrule
\end{tabular}
\end{subtable}
\begin{subtable}[h]{0.45\linewidth}
\caption{\small{Using Kinetics-pretrained feature}}
\centering
\setlength{\tabcolsep}{1.0em}
\begin{tabular}{ll}
\toprule\noalign{\smallskip}
Method & mAP  \\
\noalign{\smallskip}
\cmidrule{1-2}
\noalign{\smallskip}
IDN~\cite{eun2020idn}      & 60.3 \\
TRN~\cite{xu2019trn}       & 62.1 \\
OadTR~\cite{wang2021oadtr} & 65.2 \\
LSTR~\cite{xu2021lstr}     & 69.5 \\
Ours$^\lightning$          & {\underline{67.3}} \\
Ours                       & {\bf 71.2} \\
\bottomrule
\end{tabular}
\end{subtable}
\end{table}

\begin{table}[t]
\centering
\caption{\small{Result of online action anticipation on THUMOS'14.
$^\dagger$ was reproduced by us because LSTR~\cite{xu2021lstr} only reported ActivityNet-pretrained results}}
\label{table:online_anticipation_thumos}
\setlength{\tabcolsep}{4pt}
\begin{tabular}{lcccccccccc}
\toprule\noalign{\smallskip}
\multirow{2}{*}{method} & \multirow{2}{*}{Pre-train} & \multicolumn{8}{c}{mAP@$\tau_o$} & \multirow{2}{*}{average}   \\  \cmidrule(lr){3-10}
& & 0.25 & 0.50 & 0.75 & 1.0 & 1.25 & 1.50 & 1.75 & 2.0 & \\
\noalign{\smallskip}
\cmidrule{1-11}
\noalign{\smallskip}
RED~\cite{gao2017red} & \multirow{4}{*}{ANet1.3} & 45.3 & 42.1 & 39.6 & 37.5 & 35.8 & 34.4 & 33.2 & 32.1 & 37.5 \\
TRN~\cite{xu2019trn}       &  & 45.1 & 42.4 & 40.7 & 39.1 & 37.7 & 36.4 & 35.3 & 34.3 & 38.9 \\
TTM~\cite{wang2021oadtr}   &  & 45.9 & 43.7 & 42.4 & 41.0 & 39.9 & 39.4 & 37.9 & 37.3 & 40.9 \\
LSTR~\cite{xu2021lstr}     &  & - & - & - & - & - & - & - & - & 50.1 \\
Ours                       &  & 64.7 & 61.8 & 58.7 & 55.7 & 53.2 & 51.1 & 49.2 & 47.8 & {\bf 55.3} \\
\cmidrule{1-11}
TTM~\cite{wang2021oadtr}   & \multirow{3}{*}{K400} & 46.8 & 45.5 & 44.6 & 43.6 & 41.9 & 41.1 & 40.4 & 38.7 & 42.8 \\
LSTR$^\dagger$~\cite{xu2021lstr}     &  & 60.4 & 58.6 & 56.0 & 53.3 & 50.9 & 48.9 & 47.1 & 45.7 & 52.6 \\
Ours                       &  & 66.2 & 63.5 & 60.5 & 57.4 & 54.8 & 52.6 & 50.5 & 48.9 & {\bf 56.8}  \\ 
\bottomrule
\end{tabular}
\end{table}

\myparagraph{THUMOS'14.}
We conduct both online action detection and anticipation experiments on THUMOS'14.
In both tasks, the backbone network from which the feature is extracted is pretrained on either ActivityNet v1.3~\cite{caba2015activitynet} or Kinetics~\cite{carreira2017kinetics}.
Table~\ref{table:online_detection_thumos} shows the results of online action detection.
\testra surpasses the previous states-of-the-art by a large margin.
We also adopt NVIDIA Optical Flow SDK (NVOFA)~\footnote{\url{https://developer.nvidia.com/opticalflow-sdk}} for faster optical flow computation.
NVOFA can run as fast as 1K FPS on a $240\times180$ image sequence on a modern GPU.
We denote the model that takes NVOFA optical flow as input to be \testra$^\lightning$.
We observe some performance drop, but an mAP of 67.3\% is still competitive.
Most importantly, the runtime of the entire pipeline is significantly sped up.
More detailed discussion on runtime analysis will be provided in Sec.~\ref{sec:exp:runtime}.
All results use ES-Attention.

Table~\ref{table:online_anticipation_thumos} shows the results of online action anticipation.
\testra with Kinetics-pretrained feature achieves an average mAP of 56.8\%, outperforming all previous methods.
For fair comparison, we also rerun LSTR~\cite{xu2021lstr} using the same Kinetics-pretrained feature.
This improved LSTR is still 4\% below \testra.

\begin{table}[t]
\centering
\caption{\small{Result of action anticipation on EK100.
The upper half lists RGB-only methods; in lower half all types of inputs are allowed}}
\label{table:anticipation_ek100}
\setlength{\tabcolsep}{0.1em}
\begin{tabular}{lcccccccccccc}
\toprule %
\multirow{2}{*}{Method} & \multirow{2}{*}{Input} & \multirow{2}{*}{Pre-train} & \multicolumn{3}{c}{overall} & \multicolumn{3}{c}{unseen} & \multicolumn{3}{c}{tail} \\
\cmidrule(lr){4-6}\cmidrule(lr){7-9}\cmidrule(lr){10-12}
 &  &  & verb   & noun & action   & verb   & noun   & action   & verb  & noun  & action   \\
 \cmidrule{1-12}
 RULSTM~\cite{furnari2020rulstm} & \multirow{4}{*}{RGB} & IN-1k & 27.5 & 29.0 & 13.3 & 29.8 & 23.8 & 13.1 & 19.9 & 21.4 & 10.6 \\
 AVT~\cite{girdhar2021avt} & & IN-1k & 27.2 & 30.7 & 13.6 & - & - & - & - & - & - \\
 AVT~\cite{girdhar2021avt} & & IN-21k & 30.2 & 31.7 & 14.9 & - & - & - & - & - & - \\
 Ours & & IN-1k & 26.8 & 36.2 & 17.0 & 27.1 & 30.1 & 13.3 & 19.3 & 28.6  & 13.7  \\
\midrule
 RULSTM~\cite{furnari2020rulstm} & \multirow{4}{*}{\shortstack[c]{RGB\\+OF\\+Obj}} & IN-1k & 27.8 & 30.8 & 14.0 & 28.8 & 27.2 & 14.2 & 19.8 & 22.0 & 11.1 \\
 TempAgg~\cite{sener2020TempAgg} & & IN-1k &  23.2 & 31.4 & 14.7 & 28.0 & 26.2 & 14.5 & 14.5 & 22.5 & 11.8 \\
 AVT+~\cite{girdhar2021avt} & & IN-1k & 25.5 & 31.8 & 14.8 & 25.5 & 23.6 & 11.5 & 18.5 & 25.8 & 12.6 \\
 AVT+~\cite{girdhar2021avt} & & IN-21k & 28.2 & 32.0 & 15.9 & 29.5 & 23.9 & 11.9 & 21.1 & 25.8 & 14.1 \\
 \arrayrulecolor{black!10}\cmidrule{1-2}\arrayrulecolor{black}
 Ours & RGB+OF & IN-1k & 30.8  & 35.8 & 17.6 & 29.6 & 26.0 & 12.8 & 23.2 & 29.2 & 14.2 \\
\bottomrule
\end{tabular}
\end{table}

\myparagraph{EK100.}
We compare \testra with prior works on the EPIC-Kitchen-100 action anticipation track~\cite{Damen2021Rescaling} in Table~\ref{table:anticipation_ek100}.
We split the results into two halves: the upper half contains methods with only RGB inputs and the lower half uses additional information, such as optical flow and object feature.
Using the same ImageNet-1k-pretrained feature, \testra significantly outperforms RULSTM~\cite{furnari2020rulstm} and AVT~\cite{girdhar2021avt} on the action-level recall.
The improvement is most pronounced in the increase noun-level recall.
This demonstrates the effectiveness of incorporating longer input for anticipation.
The long-memory recalls many objects that appeared previously.
\testra with RGB+OF achieves 4.4\% higher verb-level recall than \testra with only RGB.
One reason for this our early-fusion.
Unlike late-fusion approaches, RULSTM and AVT+, we concatenate RGB and optical-flow feature at the beginning so that motion-related feature can be more effectively leveraged.
Again, all results use ES-Attention.

\subsection{Ablation Studies}
We conduct ablation experiments on EK100 to study the role of each module in the architecture.
Our full ablations uses the RGB-only model, but conclusions generally hold for two-stream input as well.

\begin{table}[t]
\centering
\caption{\small{Temporal smoothing kernels. Using explicit windowed-attention and stream-attention under the Laplace kernel yield consistent results}}
\label{table:ablation:attn}
\setlength{\tabcolsep}{0.5em}
\begin{tabular}{clccc}
\toprule %
\multirow{2}{*}{Kernel Type} & \multirow{2}{*}{Test Mode} & overall & unseen & tail \\
& & act. rec. & act. rec. & act. rec. \\
\cmidrule{1-5}
Box & FIFO (Eq.~\eqref{eq:fifo_attn_recursion})   & 16.14 & 12.64 & 12.89 \\
Box & ES (Eq.~\eqref{eq:es_attn_recursion}; $\lambda=0$) & 16.08 & 12.22 & 12.70 \\
\cmidrule{1-5}
Laplace   & ES (Eq.~\eqref{eq:es_attn_window}) & 16.95 & 13.33 & 13.73 \\
Laplace   & ES (Eq.~\eqref{eq:es_attn_recursion})   & 16.94 & 13.28 & 13.72 \\
\bottomrule
\end{tabular}
\end{table}

\myparagraph{Temporal Smoothing Kernels.}
We first verify the correctness of the temporal smoothing kernels at inference time in Table~\ref{table:ablation:attn}.
If we apply the box kernel and apply the FIFO recursion defined in Eq.~\eqref{eq:fifo_attn_recursion}, the result is 16.14\%.
However, if we use the exponential smoothing recursion defined in Eq.~\eqref{eq:es_attn_recursion} with decay factor $\lambda=0$, action recall drops by $0.2\sim0.4\%$ on unseen and tail classes.
This indicates the necessity to cache historic elements and pop them when the queue becomes full.
When using the Laplace kernel, we compare batch mode where windowed attention is computed using Eq.~\eqref{eq:es_attn_window} and stream mode where exponential smoothing recursion is computed using Eq.~\eqref{eq:es_attn_recursion}.
The results are consistent (less than 0.05\% difference).

\begin{table}[t]
\centering
\caption{\small{Ablation studies on position embeddings. Temporal position embeddings are unnecessary for long-term memory, justifying our design of separate the temporal smoothing kernel and feature vector}}
\label{table:ablation:pe}
\setlength{\tabcolsep}{0.5em}
\begin{tabular}{cc|c}
\toprule
PE @ long memory & PE @ short memory & overall act. rec. \\
\cmidrule{1-3} 
\xmark & \cmark & 17.0 \\
\cmark & \cmark & 16.8 \\
\xmark & \xmark & 15.7 \\
\bottomrule
\end{tabular}
\end{table}

\myparagraph{Effectiveness of Positional Embedding.}
The rationale behind streaming attention is that we can separate the attention kernel into temporal and feature components.
To justify this, we add a temporal positional embedding in the long-term memory and observe no performance improvement from Table~\ref{table:ablation:pe}.
We also try to remove the temporal embedding in the short-term memory but this changes the result significantly (-1.3\%).

\begin{table}[t]
\centering
\caption{\small{Ablation studies on MixClip and long-/short-term memory fusing}}
\label{table:ablation:mixclip+fuse}
\vspace{-2mm}
\begin{subtable}[h]{0.45\linewidth}
\centering
\caption{\small{The effect of MixClip}}
\label{table:ablation:mixclip}
\vspace{-1mm}
\setlength{\tabcolsep}{0.3em}
\begin{tabular}{cccccc}
\toprule\noalign{\smallskip}
MixClip Rate      & 0    & 0.2  & 0.5  & 0.8  \\
\cmidrule{1-5}
overall v. rec. & 25.8 & 26.0 & 26.8 & 26.2 \\
overall n. rec. & 34.6 & 35.3 & 36.2 & 35.2 \\
overall act. rec. & 15.5 & 16.0 & 17.0 & 16.2 \\
\bottomrule
\newline
\end{tabular}
\end{subtable}
\hspace{1em}
\begin{subtable}[h]{0.48\linewidth}
\centering
\caption{\small{Long- and short-term memory fusion}}
\label{table:ablation:fuse}
\vspace{-1mm}
\setlength{\tabcolsep}{0.3em}
\begin{tabular}{lc}
\toprule \noalign{\smallskip}
How to fuse & overall act. rec. \\
\cmidrule{1-2}
w/o. long mem. & 16.1 \\
no fuse      & 15.9 \\
@ long mem.  & 16.0 \\
@ comp. mem. & 17.0 \\
\bottomrule
\end{tabular}
\end{subtable}
\end{table}

\myparagraph{Effectiveness of MixClip}
Table~\ref{table:ablation:mixclip} shows the effect of MixClip rate on the anticipation result.
When no MixClip is applied, the baseline drops to 15.5\% action recall.
The performance consistently improves with MixClip and achieves the best (17.0\%) at $p_{mc}=0.5$.

\myparagraph{Fusing long- and short-term memory}
Table~\ref{table:ablation:fuse} compares different ways of fusing long- and short-term memory.
The naive way is to treat long- and short-term memory separately,~\ie (1) use the \testra encoder to compress distant inputs and (2) use closer inputs as queries in the \testra decoder to attend to this compressed set of vectors.
We observe that this no-fuse approach achieves 15.9\% which is even 0.2\% lower than short-memory-only baseline, where $ N=0 $ and the \testra decoder is instantiated by self-attention.
This indicates that we might need to incorporate the relationship within the short-term memory too.
To achieve this, we try to augment long-term memory by attaching short-term memory, denoted by ``@ long mem.", but see no significant improvement.
It might be because long-term memory is much longer than the short-term one so that the short-term information is overwhelmed at the first stage of compression.
Since the memory length after compressed is in the same order as the short-term memory, we concatenate both (``@ comp. mem.") and get 17.0\% action recall, improving the naive way by 1.1\%.

\subsection{Runtime Analysis}
\label{sec:exp:runtime}

\begin{figure}[t]
\centering
\begin{minipage}{0.43\textwidth}
\centering
\includegraphics[width=0.63\textwidth]{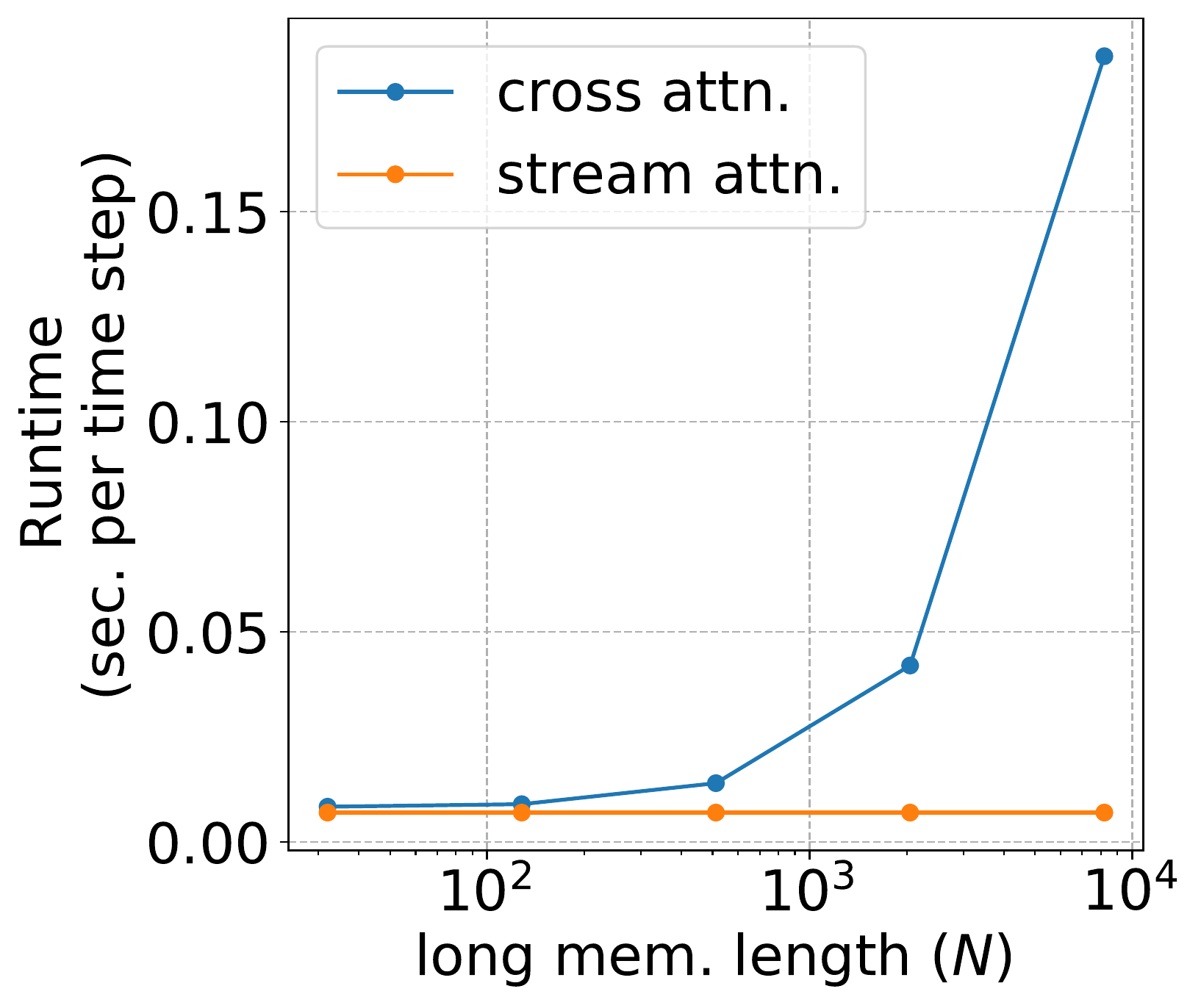}
\captionof{figure}{\small{Runtime comparison between vanilla cross attention and our exponential smoothing attention}}%
\label{fig:runtime}
\end{minipage}\hfill
\begin{minipage}{0.48\textwidth}
\centering
\setlength{\tabcolsep}{0.1em}
\begin{tabular}{lccccc}
\toprule \noalign{\smallskip}
&  OF    & RGB      & OF & \multirow{2}{*}{\testra}  & \multirow{2}{*}{Total} \\
&  Comp. & Feat. & Feat. &  \\
\cmidrule{1-6}
Ours$^\lightning$ & 1,000 & \multirow{2}{*}{150.0} & \multirow{2}{*}{104.7} & \multirow{2}{*}{142.8} & 41.1 \\
Ours & 19.3 & & & & 12.6 \\ 
\bottomrule
\end{tabular}
\captionof{table}{\small{Runtime profile (in FPS) for the entire detection system. Real-time \testra uses NVOFA optical-flow while the default one uses TV-L1~\cite{zach2007tvl1}}}%
\label{table:runtime}
\end{minipage}
\vspace{-2mm}
\end{figure}

Finally, we study the runtime speed of \testra using an NVIDIA Quadro RTX 6000 GPU.
Fig.~\ref{fig:runtime} shows the comparison of inference speed between LSTR with cross attention and \testra with ES-Attention.
We choose the length of the long memory $N$ to be $\{32, 128, 512, 2048, 8196\}$.
We can clearly see that the runtime per time step scales linearly for cross-attention-based LSTR but keeps constant for \testra.
Specifically, \testra runs at a speed of 142.8 FPS.
If we integrate \testra into the online detection system, we need to take into account of the computation overhead by the optical flow computation and feature extraction.
The runtime profile is summarized in Table~\ref{table:runtime}.
The full \testra runs at 12.6 FPS.
The bottleneck is computing optical flow using TV-L1 algorithm~\cite{zach2007tvl1}.
Using the NVOFA, the real-time \testra can run at 41.1 FPS.

\section{Conclusion}
We propose stream attention based on the kernel-based reformulation of cross-attention and apply two kinds of temporal smoothing kernels that reduce the inference computation to constant cost per frame.
The resultant temporal smoothing transformer achieves excellent performance while running at a low latency.
We hope that our design can shed some light on developing more efficient models for long-term videos understanding.

\small{
\myparagraph{Acknowledgement}
This material is in part based upon work supported by the National Science Foundation under Grant No. IIS-1845485, IIS-2006820, and the NSF Institute for Foundations of Machine Learning.
}

\clearpage
\bibliographystyle{splncs04}
\bibliography{egbib}
\end{document}